\documentclass{article}

\usepackage{microtype}
\usepackage{graphicx}
\usepackage{subfigure}
\usepackage{booktabs} 
\usepackage{soul} 
\usepackage[utf8]{inputenc} 
\usepackage[table]{xcolor} 
\usepackage{natbib} 

\usepackage{hyperref}

\usepackage[accepted]{icml2021}
\icmltitlerunning{Reconstruction of long-term historical electricity demand data}

\begin{document}

\twocolumn[
\icmltitle{Reconstruction of Long-Term Historical Electricity Demand Data
}



\icmlsetsymbol{equal}{*}

\begin{icmlauthorlist}
\icmlauthor{Reshmi Ghosh}{CMU}
\icmlauthor{Michael T. Craig}{UMich}
\icmlauthor{H. Scott Matthews}{CMU}
\icmlauthor{Constantine Samaras}{CMU}
\icmlauthor{Laure Berti-Equille}{IRD}
\end{icmlauthorlist}

\icmlaffiliation{CMU}{Carnegie Mellon University, Pittsburgh, USA}
\icmlaffiliation{UMich}{University of Michigan, Ann-Arbor, USA}
\icmlaffiliation{IRD}{Institut de Recherche pour le Développement}

\icmlcorrespondingauthor{Reshmi Ghosh}{reshmig@andrew.cmu.edu}

\icmlkeywords{Machine Learning, ICML, Deep Learning, LSTM, Time Series Modeling, Climate Change, Electricity, Power Systems, Grid Planning, Renewable Energy}

\vskip 0.3in
]



\printAffiliationsAndNotice{}  

\begin{abstract}
Long-term planning of a robust power system requires the understanding of changing demand patterns. Electricity demand is highly weather sensitive. 
Thus, the supply side variation from introducing intermittent renewable sources, juxtaposed with variable demand, will introduce additional challenges in the grid planning process. 
By understanding the spatial and temporal variability of temperature over the US, the response of demand to natural variability and climate change-related effects on temperature can be separated, especially because the effects due to the former factor are not known .Through this project, we aim to better support the technology \& policy development process for power systems by developing machine and deep learning 'back-forecasting' models to reconstruct multidecadal demand records and study the natural variability of temperature and its influence on demand. 
\end{abstract}

\section{Introduction}
\label{submission}

Inclusion of clean energy in future grid to tackle climate change requires careful and systematic planning. Both renewable energy and electricity demand have significant variability. This is because solar and wind energy are dependent on atmospheric processes 
and the seasonality of temperature is a key driver of variable demand \cite{Coker}. Peak demand in summer is not representative of the peak demand in winter months. As temperature increases, demand can be expected to increase to meet cooling needs. In regions with high electrified heating, low temperatures will also increase demand \cite{bessec}. Additionally, as the earth continues to warm, this seasonal variability in electricity demand will be further amplified due to the additional temperature gradients introduced by climate change. The effect introduced by climate change on temperature is often confounded with the natural variability of the temperature, something that is inherent to many other atmospheric factors including solar radiance, wind speed, etc. 

\textbf{Motivations:} Multi-decade historical demand data is useful to determine 'capacity gaps', that is if sufficient solar and wind resources are available during high load periods. In addition to using long time series of load data to capture the complexities involved with planning a cleaner grid, it is also useful to understand the change in electricity demand as  
residential, transportation, and commercial sectors are rapidly electrified 
Both of these analyses requires hourly historical demand records spanning multiple decades, which is missing from the Balancing Authorities (BA)\footnote{a BA is an organization in the US responsible for maintaining electricity balance within its area of operation.}database.


By assessing the spatial and temporal variability of temperature over the mainland US using machine \& deep learning, we can differentiate the effects of natural variability and climate change-related effects on temperature over demand, especially because the effects due to the former cause are lesser-known to researchers. 
 Thus, there is a need to reconstruct historical demand data with a focus to understand how natural inter-annual variability (IAV) of temperature affects demand. 




\textbf{Related Work:} Reconstruction of multi-decade historical electricity demand data as a response IAV in temperature using 'back-forecasting' methods, has not been attempted for the case of the US. 
Even in the European context, it has been discussed only in a handful of papers \cite{sp18, ths16, bloomfield1, bloomfield2, Coker}. 
The analyses in  \cite{bloomfield1, bloomfield2, Coker} studied the sensitivity of grid to climate change by using temperature data to create a 36-year long time series of demand using multiple linear regression and Generalized Additive Models for the UK, and comparing it with hourly solar and wind energy data. 

We found out that for the case of US, some studies have focused on assessing the long term IAV of wind speeds \cite{kc17, lzbh10, psb18} 
and solar radiation independently, 
but none in conjunction with corresponding electricity demand. 
Other US based studies \cite{sdlc18, rdrcl21} attempted to quantify the capacity gap in solar/wind energy during high demand spells over multiple decades (~30 years) and used demand data from a representative year due to lack of historical demand records. They replicated the representative year's data (e.g., 2018 load data \cite{rdrcl21}) for the length of period of study in their assessment. This assumption of using a single year data failed to capture variability in demand over multiple decades.
Moreover, a few studies also discussed the use Piece-wise regression to forecast demand using temperature from climate models and fixed effects (accounting for socio-economic factors) \cite{ffp, carreno2020potential}. These papers serve as a good example to develop insights about methods to forecast demand in the US, but the use of climate models for temperature records in place of reanalysis/observed temperature data fail to account for the response of demand on IAV of temperature. 

To our best knowledge, there does not exist any study that assessed hourly variability of demand and solar/wind resource in conjunction over multiple decades at a BA level, due to missing historical demand data. Thus, we aim to fill this gap by reconstructing hourly demand data between 1980-2014 for BAs by leveraging advanced regression methods and using temperature data as one of the key predictors.

\section{Proposed Method}
 
In this paper we aim to develop model architecture to reconstruct historical demand data that is generalizable (with some fine-tuning) to all BAs. We use piece-wise linear regression as suggested in \cite{bloomfield1, carreno2020potential, ffp} to 'back-forecast' demand 
and compare the results against our proposed method of Long Short Term (LSTM) architecture.
We also experimented with kernel based SVR, but it failed to perform on large dataset due to computational complexity. LSTM models have gating mechanism \cite{sorkun2020time} which helps to deal with vanishing and exploding gradients efficiently during back-propagation, and has memory state which helps in modeling dependencies in time-series data. Moreover, they have the ability to model non-linear functions. Piece-wise linear regression on the other hand is an enhanced version of linear regression, which has the capability to model non-linear dependency between dependent variables and predictors, but fails to generalize in the case of large dataset, as used in this study.
 
We are interested in conducting the analysis 
at BA level rather than national/state level because a BA is responsible for managing the electricity supply for a state/group of states (ex: ISO-NE, CAISO, NYISO, etc.). There is a need to reconstruct the demand data for all BAs individually to account for spatial heterogeneity that influence temperature. We strive to achieve maximum generalizability in our proposed model architecture and hence we use large dataset for training.  In this paper, we focus on the case study of Electric Reliability Council of Texas (ERCOT), and present our findings for one city being served by ERCOT - Dallas \footnote{we have tested our models on other locations within Texas to validate the generalizability of our proposed architecture}. 

 \textbf{Back-Forecasting:} Typically in forecasting 
the future response of a variable of interested is predicted, i.e., we try to foresee what will happen. Thus, from starting time $t$, we determine 
$t+1$, $t+2$,$\ldots$, $t+n$. In our proposed method of back-forecasting (analogous to forecasting), we leverage the same regression based methods to create historical records starting from present, i.e., from $t$, we try to predict $t-1$, $t-2$,$\ldots$, $t-n$. Hence, instead of forecasting forward in time, we forecast backward to re-create the missing data. We aim to develop these estimates for 30+ years into the past (1980 - 2014) by training on available hourly demand records between 2015-2019. 

\textbf{Data:} The available trainable data 
consists of hourly demand records (in megawatts) from ERCOT 
and hourly temperature (reported in Kelvin; converted to Celsius) from MERRA \cite{bosilovich2015merra} reanalysis \footnote{Reanalysis data is generated using data analysis methods to develop consistent records of the observed conditions, which otherwise has gaps due to techniques in which data is collected and stored} dataset between 2015-2019. We use reanalysis data as opposed to Global Climate Models as we are interested in understanding the natural IAV of temperature rather than IAV due to climate change.

The 43.8k instances of hourly demand and temperature records were divided into training and validation (hold-out) set in 80:20 ratio. 
The dependent variable (also the 'ground-truth' in validation set) is hourly demand (continuous). Hourly temperature (continuous) is one of the predictors along with fixed effects (categorical) to capture any unobserved effects which may relate to changes in temperature. The fixed effects - such as hour of the day, day of the week, month and year, were feature engineered. The hourly fixed effect captures pattern per hour of electricity consumption in different seasons, and ensures controlling for implicit behavioral patterns towards using certain appliances during different times of the day, and other decisions which are cannot be explicitly modeled otherwise. Similarly, the annual fixed effects capture any effect related to change in economic activity, population, and clean energy consumption choices, etc. Since MERRA data records for temperature only exists up to 1980, the back-forecasting of hourly demand values was only possible between 1980-2014 (test dataset timeline). 

 \textbf{Evaluation metric:} Since this is a regression based problem, we choose Root Mean Square Error (RMSE) and Goodness of Fit (adjusted $R^2$) to evaluate our models.
 
 \textbf{Modeling:} As a pre-processing step, 
 Min-max scaling was applied 
 to ensure that the models do not bias towards certain predictors. Additionally, missing values from demand data were handled by dropping them, since they only made up of less 2\% of the training+validation data. Temperature records were complete since we were using reanalysis dataset. Moreover, for Piece-wise regression, bins were created to construct segments in temperature data to ensure that the model follows different trends over different regions of the data. These bins were determined by plotting the distribution of temperature data to understand how it changed.
 
For the LSTM model, embeddings were created for categorical variables (fixed effects) since PyTorch LSTM models only accepts numeric values, and the size of the embedding was tuned during training. 
Moreover, the LSTM model expects fixed/variable length sequence to represent inputs. For the case of time series based forecasting/back-forecasting of demand data hinged on temperature as a key regressor, a fixed sequence length is more appropriate. We choose 24 as the sequence length (after multiple experiments with values between 12-36) to denote that a back-forecasted hourly demand value depends on temperature records in the past 24 hours. The sequence length parameter is an empirical choice that researchers need to make based on domain knowledge, but can also be considered as a hyper-parameter.

Both Piece-wise regression and LSTM model were deployed in python, and multiple architectures were tested for the latter, which includes stacked LSTM model, bidirectional model, etc. with P100 GPU. The  best performing model (based on RMSE loss and $R^2$ on validation set) was the LSTM based architecture consisting of a single layer of LSTM units (with dropout = 0.2) followed by two (fully connected) linear layers.  Table 1 presents the best selection of hyper-parameters which were derived after experimenting with various optimizers (Stochastic Gradient Descent \& Adam), learning rate schedulers (ReduceLRonPlateau), and learning rate values between 0.001 - 0.01.

\begin{table}[ht]
\centering
\small
\begin{tabular}{|l | l| }
\hline
Hyper-parameter type & Value \\
\hline
        Learning Rate & 0.009\\
        Epochs & 1700\\
        Optimizer & Adam\\
        Time taken per epoch & 143 sec\\
\hline

\end{tabular}
\caption{ LSTM model architecture details}
\label{tab:1}
\end{table}


\section{Results}
As hypothesized earlier, the best LSTM architecture (among all experiments)performed better than Piece-wise regression when Root Mean Squared Error (RMSE$_{piecewise}$: 1643 vs. RMSE$_{LSTM}$: 1485) and goodness of fit ($R_{piecewise}^2$: 0.73 vs. $R_{LSTM}^2$: 0.87) values were compared for Dallas (other cities within ERCOT also showed similar results). We attribute this to the complex relationship that exists between demand and temperature, which the LSTM model captures efficiently, as well as the large size of the dataset. 
 \begin{figure}[ht]
 \centering
\includegraphics[width=0.49\textwidth]{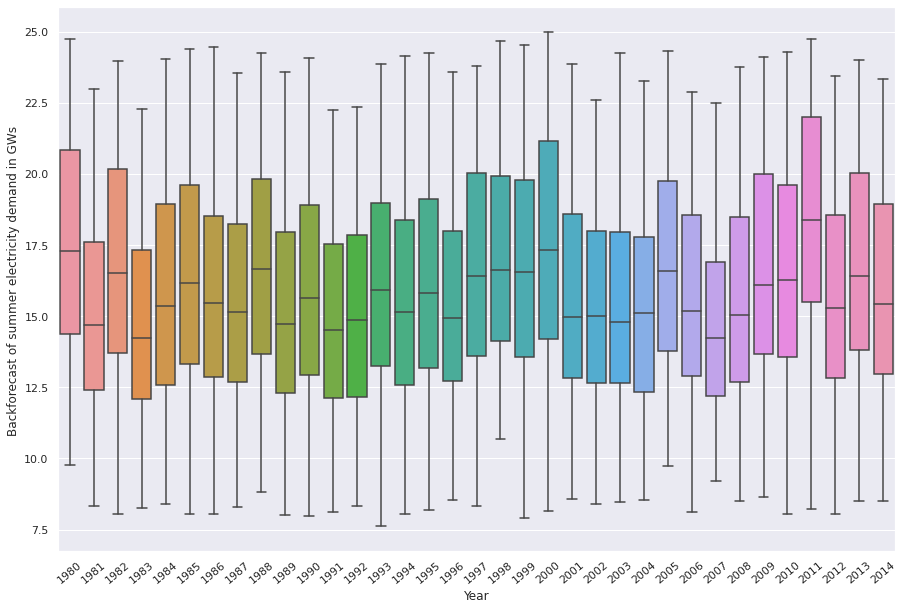}
  \caption{Distribution of back-forecasted electricity demand in GWs in the summer quarter, from 1980-2014 for Dallas}
  \label{fig:summer_ov}
  \end{figure}
  
  \begin{figure}[ht]
  \centering
    \includegraphics[width=0.49\textwidth]{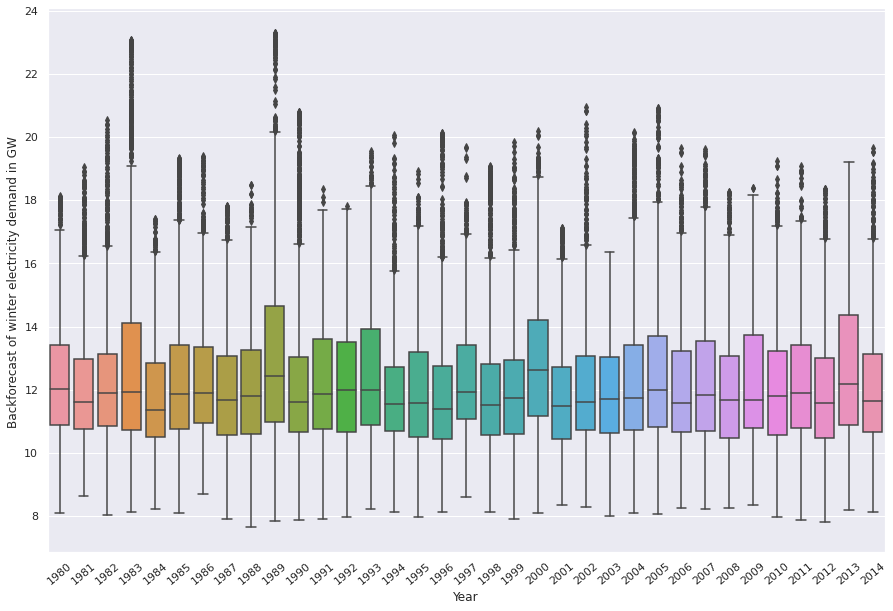}
  \caption{Distribution of back-forecasted electricity demand in GWs in the winter quarter, from 1980-2014 for Dallas}
  \label{fig:winter_ov}
  \end{figure}
 But while analysing the test set results, when we compared the distribution of back-forecasted demand in gigawatts (GWs)  grouped over the summer quarter, i.e., July - September (Figure \ref{fig:summer_ov}) and winter quarter i.e., October - December (Figure \ref{fig:winter_ov}) for Dallas, we found that the variance in winter demand was far greater than summer. This could be hypothesized to relate to sudden cold winter spells which have been common in Dallas over the years, or our proposed model was learning to back-forecast summer demand better than winter.
 
 
 We measured the $R^2$ value individually between the ground truth and predictions from summer \& winter months by slicing the backforecasts from hold-out set. We found that models were indeed performing worse on winter data ($R_{winter-LSTM}^2$: 0.48 vs. $R_{summer-LSTM}^2$: 0.83; 
 $R_{winter-piecewise}^2$: 0.21 vs. $R_{summer-piecewise}^2$: 0.78). On analyzing the scatter plot and correlation (Spearman Rank correlation coefficient -0.36) between winter temperature (Oct-Dec) for Dallas \footnote{this is true for other cities within ERCOT} and demand, we found that there is a non-monotonic relationship between the two, and the mean summer temperature is twice of winter average. Hence, a single model cannot capture the complex relationship between winter temperature and demand, and thus we decided to experiment with fitting separate models to back-forecast winter and summer demand.  
 
 
 After fitting separate LSTM models (with same architecture and hyper-parameters as the orginal) for winter and summer data, we plotted the performance on their corresponding hold-out set on an hourly basis \footnote{the 20 \% split on the total trainable data led to hold-out set only containing data from 2015. Hence, we plot maximum demand grouped on a per hour basis for winter and summer months in 2015}.

\begin{figure}[ht]
\centering
    \includegraphics[width=0.49\textwidth]{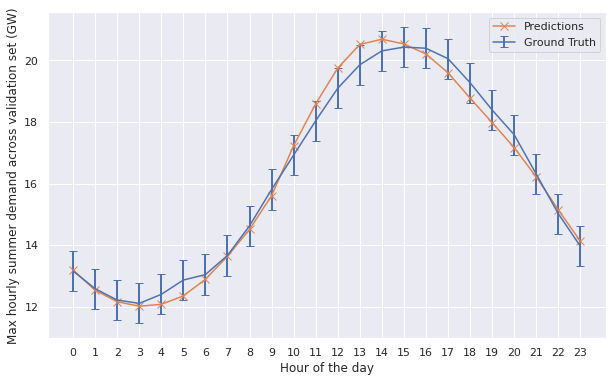}
    \caption{Maximum back-forecasted summer demand in GWs grouped per hour over validation set for Dallas}  
    \label{fig:summer_valid}
\end{figure}

\begin{figure}[ht]
\centering
    \includegraphics[width=0.49\textwidth]{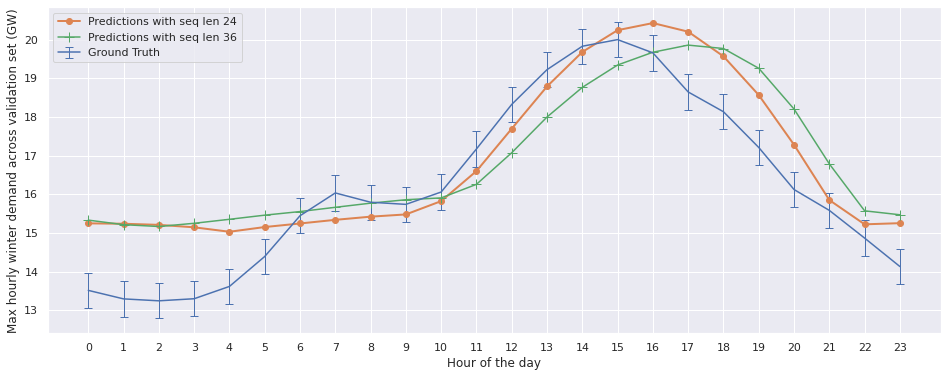}
    \caption{Maximum back-forecasted winter demand in GWs grouped per hour over validation set for Dallas}
    \label{fig:winter_valid}
\end{figure}
From Figure \ref{fig:winter_valid} and the hold-out set performance measures we found out that even though the $R_{winter}^2$ measure improved (0.76 vs. 0.48) when a separate model was used to back-forecast winter demand, it was still under performing as compared to summer demand predictions ($R_{summer}^2$ of 0.94), where almost all predictions fell in the range of $\pm$ standard error.


We attempted to change the LSTM architecture and also fine-tune the model for winter including use of different fixed sequence length as presented in Figure \ref{fig:winter_valid}, but the hold out set predictions were beyond the range of $\pm$ standard error of winter ground truth. This means tweaks in model architecture were not sufficient, \& perhaps the model needs additional weather variables as predictors (such as humidity). Additionally, between 1990 - 2016 some winters in Dallas had extreme snow spell, tornadoes, etc. Thus, additional experiments are required to identify such anomalous events and calibrating the LSTM model accordingly.

But, since the summer model performed well, we wanted to understand how ten largest hourly demand back-forecasts for Dallas evolved between 1980-2014. 
\begin{figure}[ht]
\centering
    \includegraphics[width=0.49\textwidth]{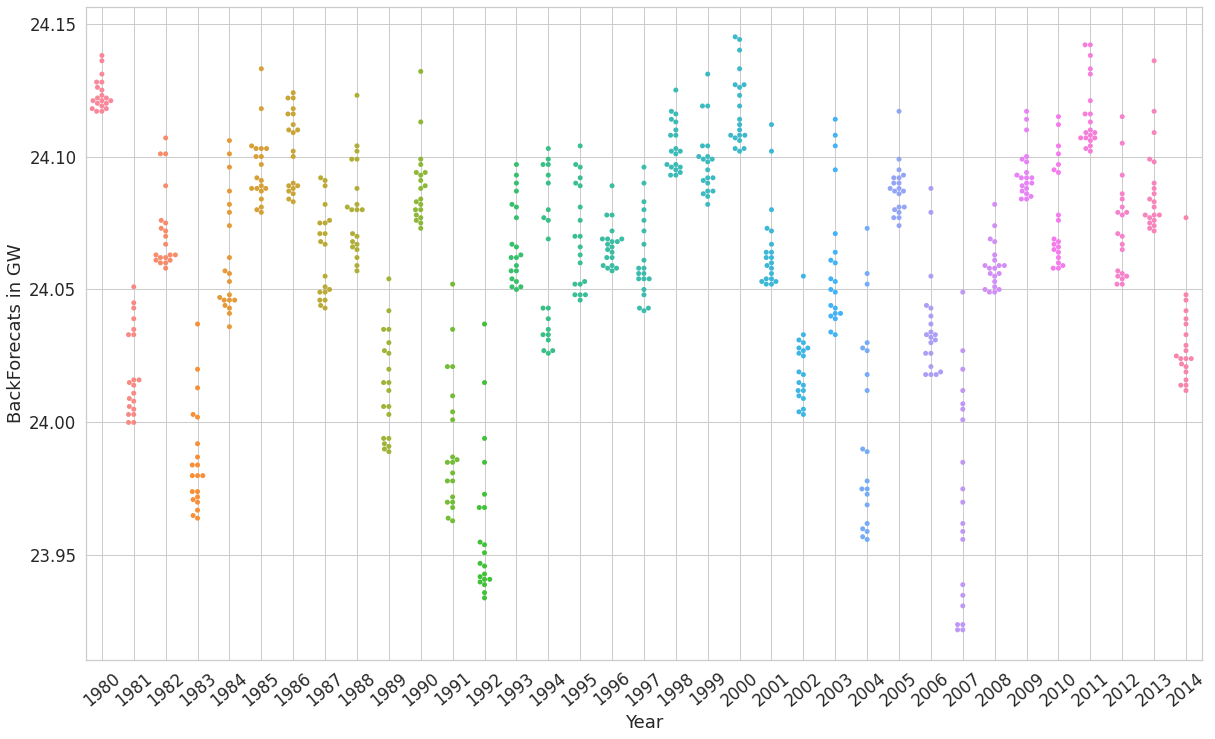}
    \caption{Twenty largest hourly demand values in GWs for Dallas between 1980-2014}
    \label{fig:summer_Dallas}
\end{figure}
The large variance in twenty greatest hourly demand requirements (Figure \ref{fig:summer_Dallas}) for 1992, 1999, 2010, 2012, etc. is evidence as to why historical demand is required to plan a grid with solar/wind resources. The available data is not representative of the past at BA level. Moreover, the trend-less' hourly demand back-forecasts corroborates the fact that there is no functional form to derive historical demand empirically.

To test the model performance in a different weather zone compared to Texas, we back-forecasted electricity demand (using the same hyper-parameters) for the North-East Massachusetts and Boston region or NEMA, governed by the Independent System Operator of New England (ISO-NE). The separate summer model performance had similar performance compared to the Dallas region ($RMSE_{summer}: 201.78$\footnote{RMSE values are scale dependent and the demand in the NEMA region of ~2GW on an hourly basis}, $R^2_{summer}: 0.80$, and $MAPE_{summer} = 6.01$). At the same time winter back-forecasts in the winter months performed relatively worse (same as the case of Dallas).

Plotting the top twenty summer demand hours in the NEMA region, we identified an interesting pattern that is significantly different than the Dallas' case. 

\begin{figure}[ht]
\centering
    \includegraphics[width=0.49\textwidth]{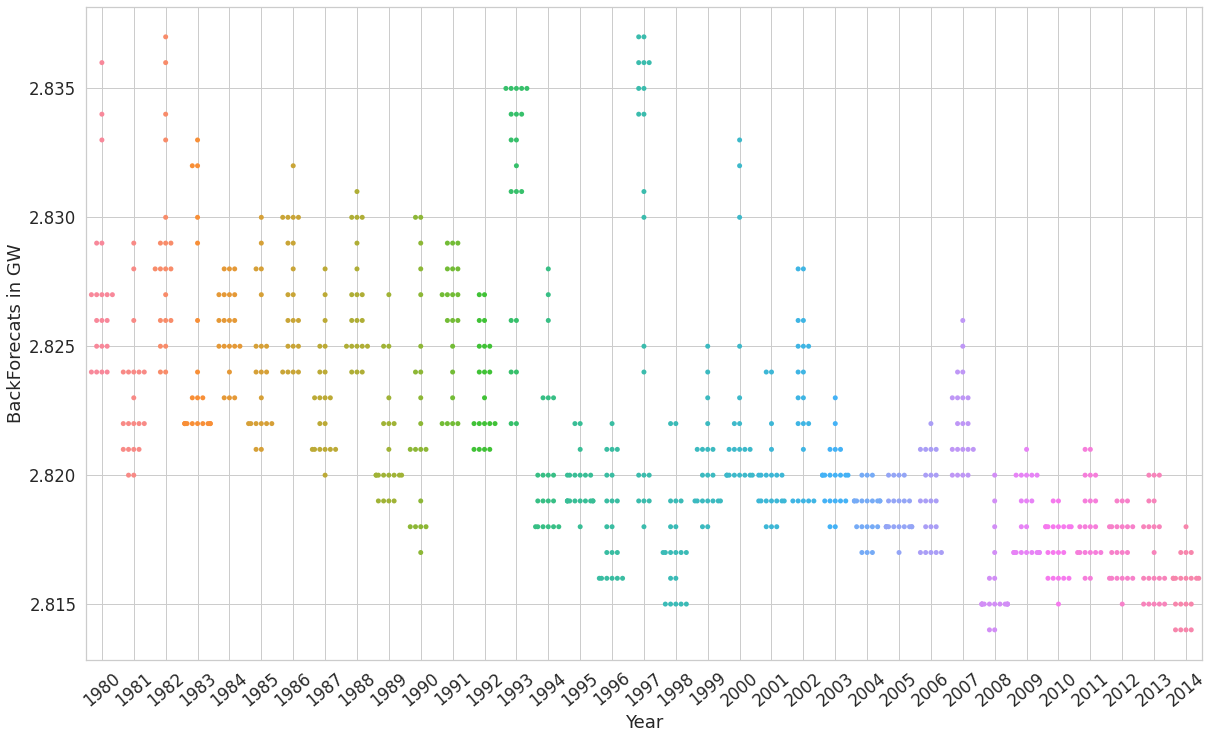}
    \caption{Twenty largest summer hourly demand values in GWs for NEMA region between 1980-2014}
    \label{fig:summer_Boston}
\end{figure}

While Dallas demand data (which falls under the North Central weather zone in ERCOT) showed large year-to-year variability in the twenty greatest summer hourly demand, the demand back-forecasts from NEMA between 1980 and 2014 showed a weak decreasing trend. This is because temperature change in different regions of the US is heterogeneous. And as these different regions experience different interannual variability in temperature, the electricity demand would change accordingly throughout the day, month, and quarter of the year. Thus, as shown while comparing the examples of ERCOT and ISO-NE, trends of electricity demand observed in one region are significantly different compared to other areas.Hence, there is a need to separately understand on a granular level the variance in hourly demand for different Balancing Authorities.    

\section{Conclusions}
In this paper, we developed a novel back-forecasting electricity demand model (reliant on natural IAV of temperature)  to reconstruct historical demand data and present a robust model to reconstruct historical summer and winter electricity demand for 30+ years for any BA (with some fine-tuning in the number of epochs \& learning rate). The model was tested on two completely different weather zones, which have different heating and cooling requirements in winter and summer respectively, that is Texas and New England. These two regions are also governed by different Balancing Authorities (ERCOT and ISO-NE), and using the performance results on validation set, as well as visualizing the distribution of back-forecasts on the test/predictor set, we concluded that a 'one model fits all' approach to back-forecasting electricity demand is not appropriate as the relationship between winter temperature and demand is more complex than summer. It is imperative to use separate models for summer and winter. Additionally, it is suggested to train models for each BA separately due to spatial heterogeneity that affects hourly temperature, although transfer learning techniques could be used.

The impact of reconstructing hourly demand data can be seen in Figure \ref{fig:summer_Dallas} and in Figure \ref{fig:summer_Boston}. It is important to capture the nuanced variability in the largest demand over different hours in a day over a long time frame, to build adequate capacity in the power system to serve all customers reliably. Aggregated annual demand data over a long time scale, even though available readily, does not help in capturing these subtle differences, which are only discernible in an hourly time frame. And thus, generating long-term hourly electricity demand data is critical to support capacity market expansion in the US.

\section{Future Work}
We firstly plan to automate the scripts used to train the models to make them "production-ready" and repeat the process of reconstructing hourly demand data for all the 66 Balancing Authorities of the US. Some of these BAs are not unique to a state, but a particular state may have multiple BAs. To account for multiple BAs in a state and not associate each one of them to the same largest population center, we plan to extract temperature carefully to correctly by using GIS data of each Balancing Authority. Furthermore, we also plan to experiment with humidity as an additional regressor in the winter LSTM model as it has been used in the demand forecasting literature and check the change in performance of the winter model. We also aim to detect anomalies in temperature data and specifically model outliers to make our suggestive winter back-forecasting architecture even more robust.

\section{Acknowledgements}
We thank Prof. Costa Samaras of Carnegie Mellon University for a being a wonderful advisor and for giving very useful suggestions. We also thank Dr. Laure Berti, for being an awesome ICML-Climate Change for AI mentor, and helping us structure the paper, while providing conference specific feedback and suggestions.



\bibliographystyle{plainnat}
\bibliography{biblography.bib}

\end{document}